\pdfoutput=1
\documentclass[11pt]{article}

\usepackage{amsmath,amsfonts,bm}

\def\eqref#1{equation~\ref{#1}}
\def\1{\bm{1}}

\DeclareMathAlphabet{\mathsfit}{\encodingdefault}{\sfdefault}{m}{sl}
\SetMathAlphabet{\mathsfit}{bold}{\encodingdefault}{\sfdefault}{bx}{n}

\usepackage[final]{acl}
\usepackage{acl}
\usepackage{authblk}
\usepackage{caption}
\usepackage{booktabs}
\usepackage[T1]{fontenc}
\usepackage{floatrow}
\usepackage{graphicx}
\usepackage{hyperref}
\usepackage[utf8]{inputenc}
\usepackage{inconsolata}
\usepackage{latexsym}
\usepackage{microtype}
\usepackage{subcaption}
\usepackage{times}
\usepackage{url}
\usepackage{xspace}
\usepackage{amsmath}
\usepackage{amsfonts}
\usepackage{mathtools}
\usepackage{listings}
\usepackage{enumitem}
\usepackage{multirow}
\usepackage{colortbl} % For row coloring
\usepackage{graphicx} % If needed for other content
\usepackage{xcolor}   % Color support
\usepackage[most]{tcolorbox}
\AtBeginEnvironment{tcolorbox}{\small}

\def\equationautorefname~#1\null{(#1)\null}
\def\itemautorefname~#1\null{(#1)\null}
\def\sectionautorefname~#1\null{\S#1\null}
\def\subsectionautorefname~#1\null{\S#1\null}
\def\subsubsectionautorefname~#1\null{\S#1\null}

\usepackage[textwidth=0.7in,disable]{todonotes}
\newcommand{\note}[4][]{\todo[author=#2,color=#3,size=\scriptsize,#1]{#4}}

\newcommand{\rafael}[2][]{\note[#1]{Rafael}{green!40}{#2}}
\newcommand{\Rafael}[2][]{\rafael[inline,#1]{#2}}

\title{Mitigating Paraphrase Attacks on Machine-Text Detectors \\ via Paraphrase Inversion}

\author[1,2]{\bf Rafael Rivera Soto}
\author[1]{\bf Barry Chen}
\author[2]{\bf Nicholas Andrews}
\affil[1]{Lawrence Livermore National Laboratory}
\affil[2]{Johns Hopkins University}
\affil[]{\texttt{rafaelriverasoto@jhu.edu, chen52@llnl.gov, noa@jhu.edu}}

\begin{document}

\maketitle

 \begin{abstract}
 High-quality paraphrases are easy to produce using instruction-tuned language models or specialized paraphrasing models.
Although this capability has a variety of benign applications, \emph{paraphrasing attacks}---paraphrases applied to machine-generated texts---are known to significantly degrade the performance of machine-text detectors.
This motivates us to consider the novel problem of paraphrase inversion, where, given paraphrased text, the objective is to recover an approximation of the original text.
The closer the approximation is to the original text, the better machine-text detectors will perform.
We propose an approach which frames the problem as translation from paraphrased text back to the original text, which requires examples of texts and corresponding paraphrases to train the \emph{inversion} model.
Fortunately, such training data can easily be generated, given a corpus of original texts and one or more paraphrasing models.
We find that language models such as GPT-4 and Llama-3 exhibit biases when paraphrasing which an inversion model can learn with a modest amount of data. 
Perhaps surprisingly, we also find that such models generalize well, including to paraphrase models unseen at training time.
Finally, we show that when combined with a paraphrased-text detector, our inversion models provide an effective defense against paraphrasing attacks, and overall our approach yields an average improvement of +22\% \texttt{AUROC} across seven machine-text detectors and three different domains.
\end{abstract}

\section{Introduction}

Recent developments in the capabilities of large language models (LLMs) such as GPT-4~\citep{openai2024gpt4technicalreport} have resulted in their widespread use by a variety of users. 
Although most users act responsibly, there is growing concern about abuses of LLMs, such as for plagiarism, spam, or spreading misinformation~\citep{taxonomyofrisks,hazell2023spearphishinglargelanguage}.
To minimize the abuse of these systems, several machine-text detection systems have been proposed, including Binoculars~\citep{hans2024spottingllmsbinocularszeroshot}, FastDetectGPT~\citep{bao2024fastdetectgptefficientzeroshotdetection}, and watermarking-based algorithms~\citep{kirchenbauer2024watermarklargelanguagemodels, kuditipudi2024robustdistortionfreewatermarkslanguage}.
However, these systems often fail to detect text that has been paraphrased by another model~\citep{krishna2020reformulatingunsupervisedstyletransfer,sadasivan2025aigeneratedtextreliablydetected}, leaving a critical gap in current detection systems.

To tackle this issue, a recent study has proposed jointly training a paraphraser and a machine-text detector with an adversarial objective: the paraphraser generates text to evade detection, while the detector identifies paraphrased text~\citep{hu2023radarrobustaitextdetection}.
Another study has proposed that LLM API providers cache their generations, enabling retrieval over a semantic space, where candidates with high similarity to previous generations are marked as paraphrases~\citep{krishna2023paraphrasingevadesdetectorsaigenerated}.
Unfortunately, both approaches lack generality, as they depend on training a specialized detector, or having access to all model generations.
A more desirable defense would be \emph{detector agnostic}, improving the performance of any detector.

Ideally, if the original tokens of a paraphrased text could be recovered, machine-text detectors would perform well, eliminating the need for any specialized solutions.
Therefore, we propose the novel task of \emph{paraphrase inversion}, where the objective is to recover the original text from a paraphrased one.
This approach has the added benefit of being detector agnostic.
Given the space of possible paraphrases and the stochastic sampling procedures commonly used, inverting paraphrased text is challenging. 
Nonetheless, there is evidence that LLMs exhibit consistent biases even when the instruction implicitly or explicitly requests diversity in the responses ~\citep{zhang2024forcing,wu2024generative}. 

Even if paraphrase inversion is possible, we must know \emph{when} to apply it, making paraphrase detection a necessary step.
Detecting text as having undergone LLM paraphrasing differs from detecting it as machine-generated, as the original text may have been human-written, in which case large portions of the original document may be copies of the human-written original.
In cases where the original text is human-written, a machine-text detector should classify it as such, for example in cases where an LLM is used as a writing assistant.

To address these concerns, we propose \emph{paraphrase detection} and \emph{paraphrase inversion} as a pipeline to improve the performance of any machine-text detector in scenarios where texts may have been paraphrased (\autoref{fig:concept_figure}). Our main contributions are as follows:

\begin{figure}[t!]
    \centering
    \includegraphics[width=1.0\linewidth]{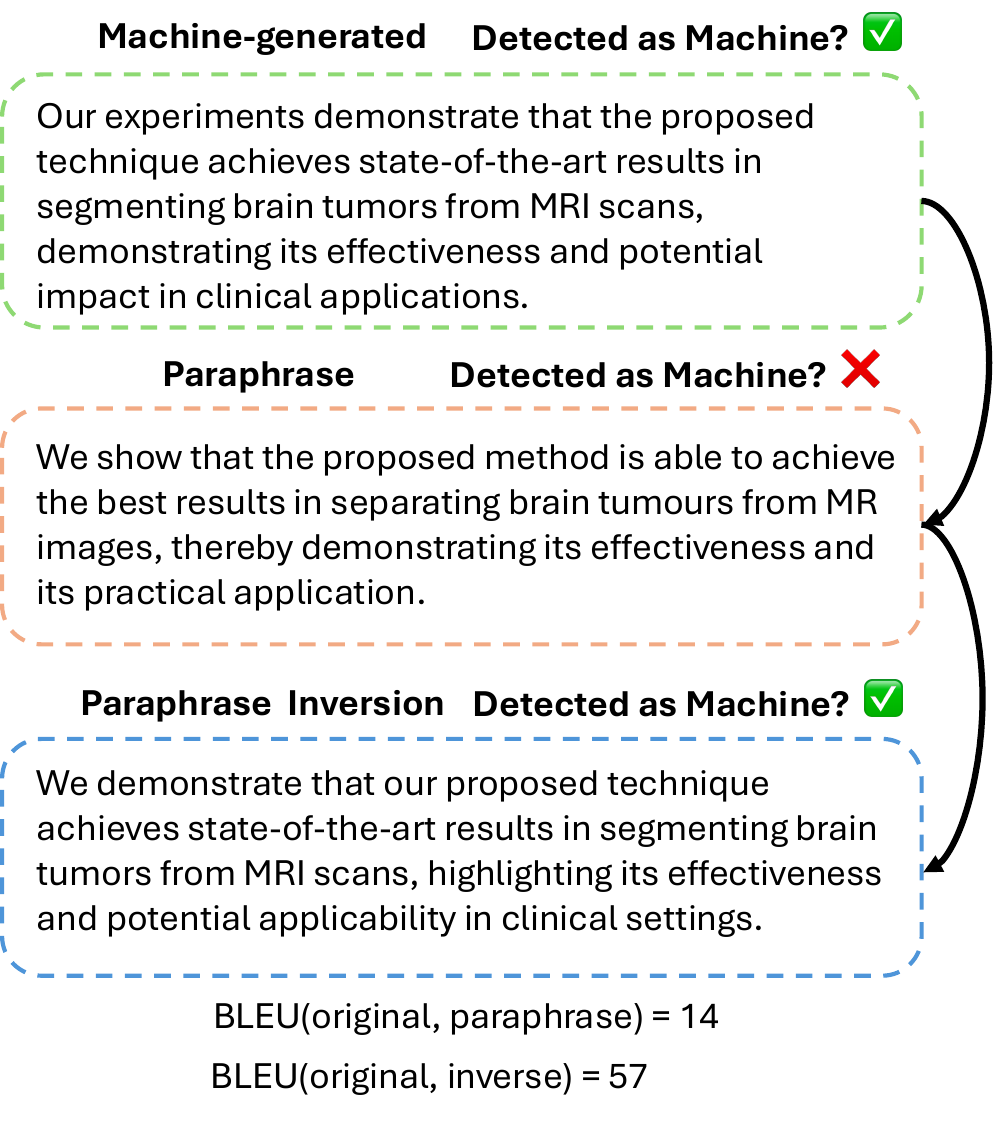}
    \caption{
    Paraphrasing defeats machine-text detection system. 
    Our proposed defense (\autoref{sec:methods}) consists of two steps: (1) detecting whether text is a paraphrase, and (2) if so, (2) inverting the paraphrase back to the original text.
    This pipeline improves the \texttt{AUROC} of 7 machine-text detectors across three domains by an average of +22\% \texttt{AUROC} (\autoref{table:machine_text_detection}).
    }
    \label{fig:concept_figure}
\end{figure}

\begin{itemize}[leftmargin=*]
    \item We introduce the task of paraphrase inversion (\autoref{sec:methods}), where the goal is to recover the original text from a paraphrased one. 
    We formalize the task and provide a comprehensive analysis of its challenges.
    We find that inverting human-written text is significantly harder than inverting paraphrases of machine-generated text, which is to be expected given that human-written text exhibits higher entropy under LLM distributions~\citep{GLTR}.
    \item We explore two paraphrase detection schemes: (1) a simple neural classifier trained to detect paraphrased text and (2) an approach that leverages  our paraphrase inversion model directly without requiring an additional model (\autoref{sec:detecting_paraphrases}). 
    \item We combine paraphrase detection and paraphrase inversion into a single pipeline that improves the detection rate of seven machine-text detectors across three domains (\autoref{sec:machine_text_detection}) by an average of +22\% \texttt{AUROC}.
\end{itemize}

\paragraph{Reproducibility} The dataset, method implementations, model checkpoints, and experimental scripts, will be released along with the paper.\footnote{Code for all experiments available \url{https://github.com/rrivera1849/inversion}}

\section{Related Work}\label{sec:related}

\paragraph{Paraphrasing} A number of paraphrase corpora have been released over the years which has enabled the development of paraphrase detection and generation models~\cite{dolan2005automatically,ganitkevitch2013ppdb,wieting-gimpel-2018-paranmt,zhang2019paws,krishna2020reformulatingunsupervisedstyletransfer}. Paraphrases have been shown to degrade the performance of machine-text detectors, including those based upon watermarking~\citep{krishna2023paraphrasingevadesdetectorsaigenerated, sadasivan2025aigeneratedtextreliablydetected}.
In response to this, several defenses have been proposed, including jointly training a paraphraser and a detector in an adversarial setting~\citep{hu2023radarrobustaitextdetection}, building  specialized detectors for both the paraphrasing model and the language model~\citep{soto2024fewshotdetectionmachinegeneratedtext}, and retrieval over a database of semantically similar generations produced by the model in the past~\citep{krishna2023paraphrasingevadesdetectorsaigenerated}.
Paraphrases have also been shown to be an effective attack against authorship verification systems~\citep{Potthast2016AuthorOA, wang2023authorshiprepresentationlearningcapture}, allowing bad actors to conceal their identity.
To our knowledge, our approach is the first attempt at inverting the paraphrases, both in general and in the context of defending against paraphrasing attacks on machine-text detection.

\paragraph{Embedding inversion} Several lines of work, both in computer vision~\citep{mahendran2014understandingdeepimagerepresentations,teterwak2021understandinginvariancefeedforwardinversion, dosovitskiy2016invertingvisualrepresentationsconvolutional} and natural language processing~\citep{song2020informationleakageembeddingmodels,li2023sentenceembeddingleaksinformation,morris2023textembeddingsrevealalmost} have explored whether embeddings can be inverted back to their inputs. 
Prior work has shown that it is possible to recover 92\% of 32-token text inputs given semantic embeddings~\citep{morris2023languagemodelinversion}.
Moreover, even when the text is isn't recovered with high-fidelity, sensitive attributes such as the authorship are recoverable~\citep{song2020informationleakageembeddingmodels}.
In computer vision, even when an inversion model is applied to an adversarially robust classifier, enough local and global detail remains, making the inversion confusable with the original image, highlighting the difficulty of safeguarding sensitive attributes~\citep{teterwak2021understandinginvariancefeedforwardinversion}.
Inverting embeddings is significantly easier than inverting paraphrases, as embeddings encode rich features of their inputs in \emph{continuous} latent-space, in contrast to the \emph{discrete} space of paraphrased tokens. 

\paragraph{Language model inversion}~\citep{morris2023languagemodelinversion} The objective here is to recover the prompt that generated a particular output.
Language model inversion techniques such as \texttt{logit2text}~\citep{morris2023languagemodelinversion} require knowledge of the LLM that generated the output \textit{and} access to the next-token probability distribution, making it difficult to apply in practice.
Another approach more closely related to ours is \texttt{output2prompt}~\citep{zhang2024extractingpromptsinvertingllm}, which trains an encoder-decoder architecture to generate the prompt given \textit{multiple} outputs.
However, \texttt{output2prompt} requires upwards of $16$ outputs per prompt to successfully match the performance of \texttt{logit2text}, and only handles prompts up to $64$ tokens long.
In contrast to these methods, we focus exclusively on inverting LLM-generated paraphrases given a \textit{single} example \textit{cleaned} of all obvious generation artifacts such as  ``\texttt{note: I changed...}", thereby removing all telltale signs of what the original text might've been. 
Therefore, the paraphrase inversion problem considered in this paper is more challenging than related problems posed in prior work. 

\section{Methods}\label{sec:methods}

\subsection{Overview}

Given a text sample $y_i$, we first detect whether it is a paraphrase using one of our detection schemes.
If it is classified as a paraphrase, we apply our paraphrase inversion model to recover the original text $\hat{x} \sim p(. \mid y_i)$.
This sample is then run through a machine-text detector.

\paragraph{Paraphrase inversion} The task of reconstructing the original source text given paraphrased text. The difficulty of this task hinges in large part on assumptions regarding the paraphrasing model.
We assume access to one or more paraphrasing models from which we can generate new paraphrases $\{y_i\}_{i=1}^N$ given a corpus of $N$ source documents $\{x_i\}_{i=1}^N$.
While access to the paraphrasing models in principle affords the possibility of producing an arbitrary amount of training data, in practice the paraphraser may be associated with non-trivial inference costs (e.g., GPT-4). 
Moreover, even if the paraphrasing model is known, the decoding parameters such as temperature may not be.\footnote{We investigate the impact of varying sampling the temperature during training and inference in~\autoref{sec:ablation}.}
Therefore, a key question is whether paraphrase inversion models generalize to \emph{unseen} paraphrasers, which we consider in~\autoref{sec:novel_inversion}.

\paragraph{Paraphrase detection} The goal is to identify whether a given text is the output of an LLM paraphraser, \emph{regardless of whether original text was human-written or machine-generated}.
Paraphrase detection is crucial for machine-text detection in the wild, where determining \emph{when} to apply a paraphrase inversion model is necessary.
We emphasize that detecting text as a paraphrase is not the same as identifying text as machine-generated, as the original text may have been human-written.
In cases where the original text is human-written, a machine-text detector should classify it as such.
This highlights the need of applying a paraphrase inversion model to ensure correct detection.
However, such a pipeline raises the risk of propagation of errors, and we should therefore carefully consider the cost of such errors.
\setlist{nolistsep}
\begin{enumerate}[noitemsep,leftmargin=*]
\item A false positive occurs when a non-paraphrased text is misidentified as paraphrased. To minimize the impact of such errors, a robust paraphrase inversion model should make \emph{minimal changes} to the text in such cases. We find that our models make significantly fewer changes to non-paraphrased documents (\autoref{sec:detecting_paraphrases}), and that this can in fact be used as a way to distinguish between paraphrased and non-paraphrased text.
\item A false negative occurs when a paraphrased text is missed by the detector and we fail to apply the inversion model. In this case, the machine-text detector is applied to the unmodified paraphrased text, which if the original text was machine-generated, is likely to result in falsely predicting that it is human written.
\end{enumerate}
Given the above considerations, the paraphrase text detector should aim for high recall at the cost of potentially lower precision. 

\subsection{End-to-end paraphrase inversion}\label{sec:learn}

\paragraph{Training objective}
The inversion models considered in this paper are fine-tuned using the standard supervised text-to-text objective, fitting an autoregressive conditional language model $p_{\theta}(y_i \mid x_i)$ on the basis of observed pairs of texts and their paraphrases $(x_i, y_i)$. Our datasets are described in~\autoref{sec:datasets}.  We parameterize all our inversion models using Mistral-7B\footnote{\texttt{\small mistralai/Mistral-7B-Instruct-v0.3}}, training it with the hyper-parameters shown in~\autoref{sec:training_hyperparameters}.
We use teacher forcing during training, conditioning on the the true observed tokens.

 \paragraph{Inference} However good the paraphrasing model, there may be considerable uncertainty in the distribution over the original text. Therefore we sample several inversions and use a scoring function to  select a single sample which scores highest.

\paragraph{Choice of score} 
A number of criteria could be optimized to help select a single inversion likely to be close to the original text. 
For example, inversions should retain the meaning of the paraphrased text, and so the score could include a measure of semantic similarity. 
Furthermore, the inversion should  be stylistically distinct from the paraphrased text, as this would indicate a return to the original machine or human styles which are known to be distinct~\cite{soto2024fewshotdetectionmachinegeneratedtext}. 
In preliminary experiments, we found that the paraphrasing model consistently preserved meaning in generated samples, and so to avoid introducing additional hyper-parameters and computational expense, we focus on stylistic distinctness. 
Specifically, we compute a stylistic embedding of the samples and original text to compute a stylistic distance for each candidate inversion, and select the inversion which is \emph{furthest}---the most stylistically distinct.
 
\subsection{Detecting Paraphrases}\label{sec:detecting_paraphrases}

\paragraph{Neural paraphrase detector} In the simplest case, we train a paraphrase detector 
 $d_\phi(. \mid y_i)$ using the standard binary-cross-entropy classification loss.
 In addition to the standard loss, we optimize the model for a paraphrased token prediction task, where the goal is to determine whether each token in a document  is copied from the original text or paraphrased.
 We include this loss to help the model capture the biases that paraphrasers introduce when rewording text.
We optimize the binary-cross-entropy for each token, corresponding to independent classification decisions.
Our model is initialized from \texttt{RoBERTa-large}\footnote{\texttt{\small FacebookAI/roberta-large}}~\citep{liu2019robertarobustlyoptimizedbert}, with a multi-layer-perceptron (MLP) head that predicts whether each token was copied from the original text or paraphrased.

\begin{figure}[t!]
    \centering
    \includegraphics[width=1.0\linewidth]{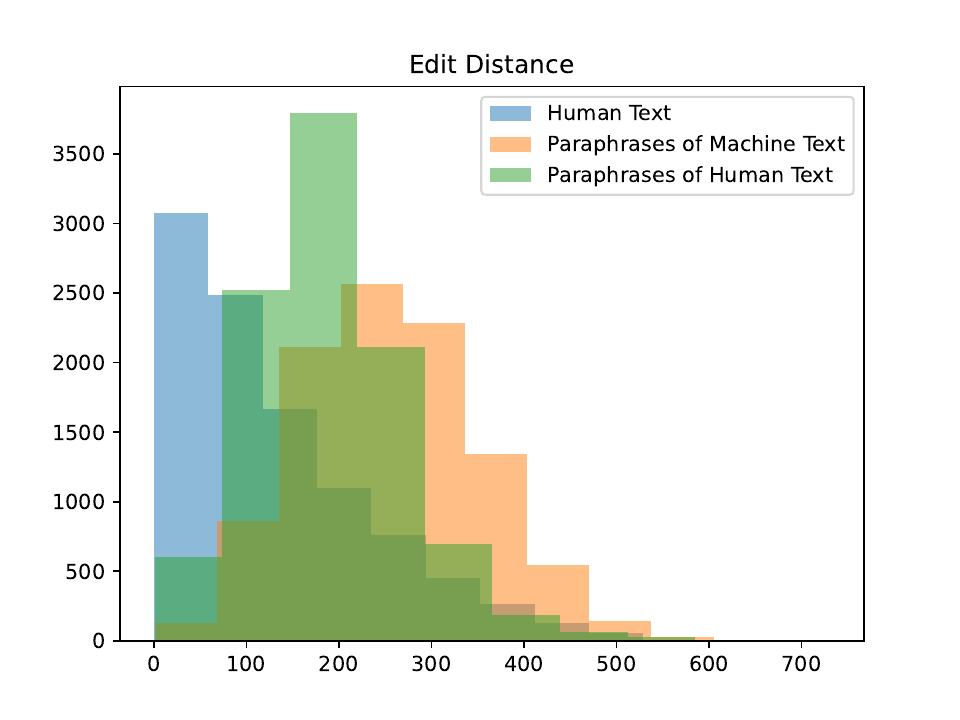}
    \caption{
    Edit distances between the original text and its inversion when the machine-paraphrase inversion model is applied to human-text and paraphrases of human- or machine-text.
    The inversion model edits human-written significantly less.
    }
    \label{fig:edit_distance}
\end{figure}

\paragraph{Edit-based paraphrase detector} Rather than training a neural classifier, we determine whether a sample $y_i$ is a paraphrase based on how many edits our paraphrase inversion model makes.
Intuitively, if the paraphrase inversion model captures LLM paraphrasing biases, it should make \emph{fewer} edits when ``inverting” a human-written text than when inverting a paraphrase.
Indeed, we find that this is the case in~\autoref{fig:edit_distance}.
This observation motivates the following paraphrase detection scheme.
Given two Gaussian distributions $g_{h}$ and $g_{m}$, where $g_{h}$ is fit on edit distances of human-text inversions and their originals and $g_{m}$ on those from paraphrases of human- and machine-text and their inversions, we detect whether a sample $y_i$ is a paraphrase by calculating whether $y_i$ is more probable under $g_{m}$ than $g_{h}$.
This is equivalent to applying a likelihood-ratio test with a threshold of $1$.
In practice, because we have $N$ inversions per sample, we take the majority vote of all such predictions.

\section{Experimental Procedure}\label{sec:experiments}

\subsection{Datasets}\label{sec:datasets}

We evaluate our approach on three domains: \texttt{Reddit}, \texttt{ArXiv}, and \texttt{MovieReviews}.
We use \texttt{Reddit} specifically to test the \emph{feasibility} of paraphrase inversion, while all three domains are used to evaluate our pipeline for defending machine-text detectors against paraphrase attacks.
The validation sets of each domain are used to train our edit-based paraphrase detector introduced in~\autoref{sec:detecting_paraphrases}, while the training sets are used to train our paraphrase inversion models and our neural paraphrase detector.
The \texttt{ArXiv} and \texttt{MovieReviews} datasets are subsampled from the RAID~\citep{dugan2024raidsharedbenchmarkrobust} dataset, a machine-text detection benchmark which contains paraphrases of machine-text using DIPPER~\citep{krishna2023paraphrasingevadesdetectorsaigenerated}.
We refer to these two datasets as \texttt{RAID-ArXiv} and \texttt{RAID-MovieReviews}.
The details of how RAID was subsampled can be found in~\autoref{sec:raid_creating}.
Here, we discuss how we generate human-text paraphrases and machine-text paraphrases for the \texttt{Reddit} domain, as well as the construction of the \texttt{Reddit} machine-detection dataset.

\paragraph{Human-text paraphrases}
We use the Reddit Million User Dataset (MUD), which contains comments from over 1 million Reddit users over a wide variety of topics~\citep{khan2021deepmetriclearningapproach}. 
We subsample the dataset according to the procedure in~\autoref{sec:subsampling_reddit}.
Once subsampled, we generate the paraphrases of human-text by prompting Mistral-7B\footnote{\texttt{\small mistralai/Mistral-7B-Instruct-v0.3}}~\citep{jiang2023mistral7b}, Phi-3B\footnote{\texttt{\small microsoft/Phi-3-mini-4k-instruct}}~\citep{abdin2024phi3technicalreporthighly}, and Llama-3.1-8B\footnote{\texttt{\small meta-llama/Meta-Llama-3-8B-Instruct}}~\citep{dubey2024llama3herdmodels}.
We clean all obvious LLM-generated artifacts such as \texttt{This rephrased passage condenses}, \texttt{note: I changed...}, and ensure that all paraphrases have a semantic similarity of at least 0.7 under SBERT\footnote{\texttt{sentence-transformers/all-mpnet-base-v2}}~\citep{reimers2019sentencebertsentenceembeddingsusing}.

\paragraph{Machine-text paraphrases}
To generate paraphrases of \textit{machine-text}, we first prompt one of the three LLM at random to produce a response to each human-written comment, then we follow the same paraphrasing procedure described above.

\paragraph{Machine-text detection}
We combine the test set of both our human-text paraphrase and machine-text paraphrase datasets to create a new set composed of $500$ samples in each category: human text, paraphrases of human text, and paraphrases of machine text.

\begin{table*}[htb!]
\centering\small
\begin{tabular}{lccc}
\bf Detector & \multicolumn{3}{c}{\bf AUROC} \\
& \bf Baseline & \bf Inversion+Edit-based & \bf Inversion+Neural \\\toprule
\rowcolor[gray]{0.85} \multicolumn{4}{c}{Reddit} \\
\bf OpenAI~\citeyearpar{solaiman2019releasestrategiessocialimpacts}  & 0.56 & 0.77 & \bf 0.79 \\
\bf Rank~\citeyearpar{GLTR} & 0.56 & 0.66 & \bf 0.68 \\
\bf LogRank~\citeyearpar{solaiman2019releasestrategiessocialimpacts} & 0.58 & 0.74 & \bf 0.77 \\
\bf Entropy~\citeyearpar{ippolito-etal-2020-automatic} & 0.51 & \bf 0.59 & \bf 0.59 \\
\bf RADAR~\citeyearpar{hu2023radarrobustaitextdetection} & 0.62 & 0.66 & \bf 0.70 \\
\bf FastDetectGPT~\citeyearpar{bao2024fastdetectgptefficientzeroshotdetection} & 0.66 & 0.80 & \bf 0.84 \\
\bf Binoculars~\citeyearpar{hans2024spottingllmsbinocularszeroshot} & 0.77 &  0.84 & \bf 0.89 \\
\midrule
\rowcolor[gray]{0.85} \multicolumn{4}{c}{RAID-ArXiv} \\
\bf OpenAI~\citeyearpar{solaiman2019releasestrategiessocialimpacts}  & \bf 0.81 & 0.79 & 0.77 \\
\bf Rank~\citeyearpar{GLTR} & 0.71 & 0.69 & \bf \bf 0.79 \\
\bf LogRank~\citeyearpar{solaiman2019releasestrategiessocialimpacts} & 0.75 & 0.72 & \bf \bf 0.91 \\
\bf Entropy~\citeyearpar{ippolito-etal-2020-automatic} & 0.39 & 0.42 & \bf 0.62 \\
\bf RADAR~\citeyearpar{hu2023radarrobustaitextdetection} & \bf 0.99 & 0.98 & \bf 0.99 \\
\bf FastDetectGPT~\citeyearpar{bao2024fastdetectgptefficientzeroshotdetection} & 0.83 & 0.78 & \bf 0.91 \\
\bf Binoculars~\citeyearpar{hans2024spottingllmsbinocularszeroshot} & 0.92 & 0.86 & \bf 0.98 \\
\midrule
\rowcolor[gray]{0.85} \multicolumn{4}{c}{RAID-MovieReviews} \\
\bf OpenAI~\citeyearpar{solaiman2019releasestrategiessocialimpacts}  & 0.82 & 0.77 & \bf 0.83 \\
\bf Rank~\citeyearpar{GLTR} & 0.60 & 0.76 & \bf 0.84 \\
\bf LogRank~\citeyearpar{solaiman2019releasestrategiessocialimpacts} & 0.66 & 0.84 & \bf 0.91 \\
\bf Entropy~\citeyearpar{ippolito-etal-2020-automatic} & 0.39 & 0.63 & \bf 0.71 \\
\bf RADAR~\citeyearpar{hu2023radarrobustaitextdetection} & 0.92 & 0.92 & \bf 0.95 \\
\bf FastDetectGPT~\citeyearpar{bao2024fastdetectgptefficientzeroshotdetection} & 0.74 & 0.80 & \bf 0.89 \\
\bf Binoculars~\citeyearpar{hans2024spottingllmsbinocularszeroshot} & 0.91 & 0.92 & \bf 0.96 \\
\midrule
\end{tabular}
\Rafael{Consider adding ``oracle" -- assuming perfect knowledge of what has been paraphrased.}
\caption{
Machine-text detection performance on a dataset of human-text, paraphrases of human-text, and paraphrases of machine-text.
Applying our inversion model to all samples detected as paraphrases using our paraphrase detection schemes~(\autoref{sec:detecting_paraphrases}), we observe significant improvements in detection performance.
}
\label{table:machine_text_detection}
\end{table*}

\subsection{Metrics}\label{sec:metrics}

To measure how well the inverted text recovers the true tokens, we make use of BLEU~\citep{papineni-etal-2002-bleu}, a measure of n-gram overlap.
Recovering the original tokens may be difficult, if not impossible.
As such, we posit that the inverted text should be close both in style and semantics to the original.
We measure the stylistic similarity by embedding the inversion and the original using \texttt{LUAR}~\citep{rivera-soto-etal-2021-learning}\footnote{\texttt{\small rrivera1849/LUAR-CRUD}}, a model that captures the stylistic features of text; we report the stylistic similarity as the cosine similarity between the embeddings.
For semantic similarity, we use \texttt{SBERT}~\citep{reimers2019sentencebertsentenceembeddingsusing} to embed the texts and report the cosine similarity between them.
To test the performance of the machine-text detectors, we report the area under the curve (\texttt{AUC}) of the receiver operating curve (\texttt{ROC}), here denoted as \texttt{AUROC}.

\subsection{Baselines}\label{sec:baselines}

For comparison, we prompt GPT-4 to invert the paraphrases.
We report the prompts used in~\autoref{sec:prompts_inversion}.
Additionally, we compare our inversion model to \texttt{output2prompt}~\citep{zhang2024extractingpromptsinvertingllm}, training it on the same dataset.
For machine-text detection, we avail of many popular detectors. We use Rank~\citep{GLTR}, LogRank~\citep{solaiman2019releasestrategiessocialimpacts}, Entropy~\citep{ippolito-etal-2020-automatic}, OpenAI's detector~\citep{solaiman2019releasestrategiessocialimpacts}, RADAR~\citep{hu2023radarrobustaitextdetection}, FastDetectGPT~\citep{bao2024fastdetectgptefficientzeroshotdetection}, and Binoculars~\citep{hans2024spottingllmsbinocularszeroshot}.

\section{Main Results}\label{sec:results}

This section present results for our motivating application of defending against paraphrasing attacks for machine-text detection. %in~\autoref{sec:paraphrase_detection_results} and~\autoref{sec:machine_text_detection}.
Next, in \autoref{sec:analysis}, we perform further analysis of individual components of our approach, including the feasibility of paraphrase inversion as a stand-alone task, considering both inversions of paraphrased machine-generated (\autoref{sec:inverting_machine_written}) and inversions paraphrased human-written documents (\autoref{sec:inverting_human_written}). 

\subsection{Paraphrase detection}\label{sec:paraphrase_detection_results}

We evaluate the proposed paraphrased detection schemes described in~\autoref{sec:detecting_paraphrases}.
We train the methods in all three domains, and report results in~\autoref{table:paraphrase_detection}. 
We find that the neural detector outperforms the edit-based detector across two out of three of the domains.
Moreover, the edit-based detector performs poorly in RAID-ArXiv, the most challenging domain, which in turn harms the performance of machine-text detectors in this setting (\autoref{sec:machine_text_detection}).

\begin{table}[t!]
\centering\small
\begin{tabular}{lcc}
\bf Dataset & \bf Edit-based & \bf Neural \\\toprule
\bf Reddit & 0.79 & \bf 0.94 \\
\bf RAID-ArXiv & 0.52 & \bf 0.67 \\
\bf RAID-Reviews & \bf 0.79 & 0.72 \\
\midrule
\end{tabular}
\Rafael{Add numbers.}
\caption{F1 scores for the proposed paraphrased detection schemes (\autoref{sec:detecting_paraphrases}).}
\label{table:paraphrase_detection}
\end{table}

\begin{table*}[htb!]
\centering
\small
\begin{tabular}{llccc|ccc}
\bf Method & \bf Type & \multicolumn{3}{c}{\bf Machine-written Text} & \multicolumn{3}{c}{\bf Human-written Text} \\
& & \bf Style $(\uparrow)$ & \bf Meaning $(\uparrow)$ & \bf BLEU $(\uparrow)$ & \bf Style $(\uparrow)$ & \bf Meaning $(\uparrow)$ & \bf BLEU $(\uparrow)$ \\
\toprule
\bf Paraphrases & - & 0.80 & 0.88 & 0.17 & 0.51 & 0.82 & 0.08 \\
\midrule
\rowcolor[gray]{0.85} \multicolumn{8}{c}{Baselines} \\
\bf GPT-4 & Single & 0.80 & 0.85 & 0.20 & 0.50 & 0.80 & 0.07 \\
& Max & 0.86 & 0.90 & 0.33 & 0.56 & 0.84 & 0.11 \\
& Mean & 0.80 & 0.87 & 0.21 & 0.50 & 0.80 & 0.07 \\
\bf out2prompt & Single & 0.48 & 0.17 & 0.00 & 0.39 & 0.10 & 0.00 \\
& Max & 0.71 & 0.40 & 0.04 & 0.53 & 0.32 & 0.02 \\
& Mean & 0.48 & 0.17 & 0.00 & 0.39 & 0.09 & 0.00 \\
\midrule
\rowcolor[gray]{0.85} \multicolumn{8}{c}{Ours} \\
\bf Inversion & Single & 0.84 & 0.90 & 0.34 & 0.54 & 0.81 & 0.13 \\
& Max & \bf 0.91 & \bf 0.95 & \bf 0.51 & \bf 0.70 & \bf 0.90 & \bf 0.25 \\
& Mean & 0.84 & 0.90 & 0.35 & 0.54 & 0.81 & 0.12 \\
\end{tabular}
\caption{
Results of inverting \textit{paraphrases of machine-written text} (left three columns) and \textit{paraphrases of human-written text} (right three columns). We generate $100$ inversions per sample and report the metrics achieved by a single inversion, by the best inversion (max), and the average across all inversions.
Our proposed inversion model outperforms all baselines.
}
\label{table:machine_inversion}
\end{table*}

\begin{table}[htb!]
\centering\small
\begin{tabular}{lcc}
\bf Detector & \multicolumn{2}{c}{\bf AUROC} \\
& \bf Baseline & \bf Inversion \\\toprule
\rowcolor[gray]{0.85} \multicolumn{3}{c}{Train - RAID-MovieReviews, Eval - RAID-ArXiv} \\
\bf OpenAI~\citeyearpar{solaiman2019releasestrategiessocialimpacts}  & 0.81 & \bf 0.84 \\
\bf Rank~\citeyearpar{GLTR} & 0.71 & \bf 0.83  \\
\bf LogRank~\citeyearpar{solaiman2019releasestrategiessocialimpacts} & 0.75 & \bf 0.89 \\
\bf Entropy~\citeyearpar{ippolito-etal-2020-automatic} & 0.39 & \bf 0.68 \\
\bf RADAR~\citeyearpar{hu2023radarrobustaitextdetection} & \bf 0.99 & \bf 0.99 \\
\bf FastDetectGPT~\citeyearpar{bao2024fastdetectgptefficientzeroshotdetection} & 0.83 & \bf 0.90 \\
\bf Binoculars~\citeyearpar{hans2024spottingllmsbinocularszeroshot} & 0.92 & \bf 0.96 \\
\midrule
\rowcolor[gray]{0.85} \multicolumn{3}{c}{Train - RAID-ArXiv, Eval - RAID-MovieReviews} \\
\bf OpenAI~\citeyearpar{solaiman2019releasestrategiessocialimpacts}  & \bf 0.82 & \bf 0.82 \\
\bf Rank~\citeyearpar{GLTR} & 0.60 & \bf 0.83 \\
\bf LogRank~\citeyearpar{solaiman2019releasestrategiessocialimpacts} & 0.66 & \bf 0.90 \\
\bf Entropy~\citeyearpar{ippolito-etal-2020-automatic} & 0.39 & \bf 0.68 \\
\bf RADAR~\citeyearpar{hu2023radarrobustaitextdetection} & 0.92 & \bf 0.94 \\
\bf FastDetectGPT~\citeyearpar{bao2024fastdetectgptefficientzeroshotdetection} & 0.74 & \bf 0.87 \\
\bf Binoculars~\citeyearpar{hans2024spottingllmsbinocularszeroshot} & 0.91 & \bf 0.95 \\
\midrule
\end{tabular}
\caption{
Machine-text detection performance on a dataset of human-text, paraphrases of human-text, and paraphrases of machine-text.
We find that when our pipeline generalizes even when trained on one domain, and evaluated on another (e.g. \texttt{RAID-ArXiv} $\rightarrow$ \texttt{RAID-MovieReviews}).
}
\label{table:machine_text_detection_general}
\end{table}

\subsection{Machine-Text Detection} \label{sec:machine_text_detection}

We consider the scenario where human- or machine-text may have been paraphrased by an LLM.
In this scenario, it would be desirable to label paraphrases of human-text as human-written and paraphrases of machine-text as machine-generated.
We train and evaluate our defense pipeline on all three domains separately.
We run our paraphrase detection schemes on the held-out test set, inverting each sample detected as a paraphrase $100$ times, and picking the inversion that is the \emph{farthest} away from the input-text in \texttt{LUAR} space, ensuring that the style is dissimilar from paraphrasing style.
We report the \texttt{AUROC} of 7 popular machine-text detectors in~\autoref{table:machine_text_detection}, and make the following observations:
(1) Our defense, with the neural paraphrase detector improves the performance of 7 machine-text detectors across 3 domains. 
The only exception is OpenAI's detector on the \texttt{RAID-ArXiv} dataset. 
(2) RADAR, a detector designed to be robust against paraphrase attacks, also benefits.
Indeed, in the worst case, RADAR's performance remains unchanged (\texttt{RADAR-ArXiv}), but in other domains, we observe notable improvements.
This highlights that our defense can be combined with other existing defenses.
(3) The edit-based paraphrase detector is not robust across all domains. Although the edit-based paraphrase detector improves performance on the \texttt{Reddit} and \texttt{RAID-MovieReviews} datasets, it reduces performance on \texttt{RAID-ArXiv}.
This decline is due to the many mis-classifications in that domain.
However, overall we observe an average improvement of +22\% \texttt{AUROC} averaged across all detectors and domains.

\subsection{Generalizing across domains}~\label{sec:generalization_across_domains} 
Do the paraphrase detection and paraphrase inversion models generalize from one dataset to another? 
We apply the pipeline using the neural paraphrase detector and inversion model trained on \texttt{RAID-ArXiv} to \texttt{RAID-MovieReviews}, and vice versa, showing our results in~\autoref{table:machine_text_detection_general}.
We find that our pipeline improves results across all detectors even under these conditions, suggesting that paraphrasers exhibit similar biases regardless of what domain they're applied to.

\section{Further Analysis}\label{sec:analysis}

\begin{table*}[htb!]
\centering\small
\begin{tabular}{lcccc}
\bf Method & \bf Type & \bf Style Sim. $(\uparrow)$ & \bf Semantic Sim. $(\uparrow)$ & \bf BLEU $(\uparrow)$ \\
\toprule
\bf Paraphrases & - & 0.61 & 0.90 & 0.21 \\
\midrule
\bf Inversion & Single & 0.62 & 0.88 & 0.26 \\
& Maximum & \bf 0.77 & \bf 0.94 & \bf 0.41 \\
& Average & 0.62 & 0.88 & 0.26 \\
\midrule
\end{tabular}
\caption{
Inverting GPT-4 paraphrases of human-text, an LLM \emph{unseen} by the inversion model during training time.
We generate $100$ inversions per sample, and report the metrics achieved by a single inversion, by the best inversion (maximum), and the average across all inversions.
}
\label{table:novel_inversion}
\end{table*}

\subsection{Inverting paraphrases of machine-generated text}\label{sec:inverting_machine_written}

In this section, we explore the extent to which paraphrases of \textit{machine-generated} text can be inverted to their original tokens.
We expect this task to be easier than inverting paraphrases of \textit{human-written} text, as human-written tokens exhibit high entropy under LLM distributions~\citep{GLTR}.
We train and evaluate all models on Reddit, generating $100$ inversions per sample on the held-out test set and report metrics in~\autoref{table:machine_inversion}.
We observe that our model recovers significant portions of the original text, with the best-scoring inversions achieving an average BLEU score of 51, with semantic and stylistic similarities of $0.95$ and $0.91$, respectively.

\subsection{Inverting paraphrases of human-written text}\label{sec:inverting_human_written}

We now turn to the more difficult problem of inverting paraphrases of \textit{human-written} text. 
We train and evaluate all models on Reddit, generating $100$ inversions per sample on the held-out test set and report metrics in~\autoref{table:machine_inversion}.
We highlight some key observations:
(1) Inverting paraphrases of human-written text is harder than paraphrases of machine-generated text, with the best scoring inversions achieving an average BLEU score of $25$, which is half of that achieved when inverting paraphrases of machine-written text (\autoref{sec:inverting_machine_written}).
(2) \texttt{output2prompt} does not recover significant portions of the original-text, we attribute this to its requirement of observing multiple outputs per prompt, and to the fact that the model has much lower capacity than ours (T5-base vs Mistral-7B).

\subsection{Can inversion  models invert a novel paraphraser?}\label{sec:novel_inversion}
To answer this question, we prompt GPT-4, an unseen LLM during training time, to paraphrase the human-written \texttt{Reddit} test set.
We use our inversion model trained on Reddit to invert each paraphrase $100$ times, and report the metrics in~\autoref{table:novel_inversion}.
Surprisingly, we find that \textbf{GPT-4 is easier to invert than the models seen during training}, with our model achieving a BLEU score of $41$.
We attribute this to GPT-4 paraphrases retaining more of the original text, with its paraphrases achieving a BLEU score of $21$ in contrast to the BLEU score of $8$ achieved by the LLMs used for training (\autoref{table:machine_inversion}).

\begin{table}[t!]
\centering\small
\begin{tabular}{lc}
\bf Model & \bf BLEU \\\toprule
\bf Phi-3 & 0.08 \\
\bf Mistral-7b & 0.11 \\
\bf Llama-3-8B & 0.08 \\
\midrule
\end{tabular}
\caption{LLMs prompted to invert their own paraphrases both with, and without in-context examples.
Generated $100$ inversions per sample, best BLEU score per sample shown.
}
\label{table:prompting_same}
\end{table}

\subsection{Can an LLM invert its own paraphrases?}\label{sec:LLM_inverting_own}

We prompt each LLM that generated a paraphrase in our \texttt{Reddit} dataset to invert its own paraphrase.
We generate $100$ inversions, and report the average maximum BLEU score achieved in~\autoref{table:prompting_same}.
Overall, we find when prompted, state-of-the-art LLMs are unable to invert their own paraphrase.
This implies that even if some parametric knowledge encodes the paraphrasing process, the LLM is not able to recover the original text given a paraphrase, further motivating our approach of training paraphrase inversion models.  

\section{Conclusion}\label{sec:conclusion}
\paragraph{Summary of findings} 
In this paper, we presented the first detector-agnostic defense against paraphrase attacks.
This defense relies on the novel task of \emph{paraphrase inversion}, where the goal is to recover the original tokens of paraphrased text. 
Furthermore, we proposed two paraphrase detection schemes: one based upon a neural-classifier and another that relies on the number of edits our inversion model makes.
When combined with one of the proposed paraphrase detectors, our pipeline improves the results of 7 machine-text detectors across 3 domains by an average of +22\% \texttt{AUROC}.
We attribute the effectiveness of our defense to the \emph{stylistic} similarity of the inverted paraphrases to the original text, which is sufficient for machine text detectors to accurately classify the inverted text.
Furthermore, we show that when our defense is trained on one domain, it generalizes to another, suggesting that paraphrasers exhibit consistent biases that can be exploited both for detecting paraphrased text and for learning to invert them.

\section*{Limitations} The number of paraphrases we use to train our inversion models is limited by our compute budget. We expect that training on additional LLM-generated paraphrases will improve all the results reported in the paper; as such, the results reported here should be viewed as a lower bound on achievable performance. Our compute budget also precluded experimenting with larger local models such as Llama-3 70B; however, we do include results with GPT-4 which is of comparable or greater quality.

% \bibliographystyle{acl_natlib}
% \setcitestyle{authoryear, maxbibnames=3, minbibnames=1}
\bibliography{main}

\appendix
\label{sec:appendix}

\section{Subsampling the Reddit Dataset}\label{sec:subsampling_reddit}

We subsample the dataset to authors who post in \texttt{r/politics} and \texttt{r/PoliticalDiscussion}, keeping comments composed of at least $64$ tokens but no more than $128$ tokens according to the LUAR tokenizer. 
Furthermore, we remove authors with less than $10$ comments, and randomly sample $10$ comments from all others, ensuring that no author is over-represented.

To learn to invert paraphrases, we must observe a diverse set of source documents and corresponding paraphrases. However, a random sample of documents may not provide broad enough coverage of writing styles.
For example, when we prompt GPT-4 to generate a paraphrase of "HELLO WORLD", it produces "Greetings, Universe!", removing the capital letters.
Without observing authors who write only with capital letters during training, it would be impossible for the  inversion model to invert the paraphrase. 
As such, we split authors into training, validation, and testing splits by sampling authors evenly across the \textit{stylistic} space.
We use \texttt{LUAR}~\citep{rivera-soto-etal-2021-learning}, an embedding that captures stylistic features, to embed each author's posts into a single stylistic embedding.
Then, we cluster the dataset using K-Means, setting $K=100$.
Finally, we take 80\% of the authors from each cluster for training, 10\% for validation, and randomly sample 100 authors ($2,449$ posts) of those remaining for testing.

\section{Creating Datasets from RAID}\label{sec:raid_creating}

In contrast to our \texttt{Reddit} dataset, the RAID~\citep{dugan2024raidsharedbenchmarkrobust} benchmark doesn't contain author-labels.
Therefore, sampling authors evenly across stylistic space as in~\autoref{sec:datasets} is not possible. 
RAID contains paraphrases of machine-text using DIPPER, but lacks paraphrases of human-text.
To address this, we paraphrase all human-text within \texttt{ArXiv} and \texttt{MovieReviews} with DIPPER, using the same hyper-parameters as the creators of RAID (60 lexical diversity, 0 order diversity, 512 max-tokens).
We pair up the machine-text with their corresponding paraphrases, randomly sampling 80\% of these pairs for training, 10\% for validation, and 10\% for testing.
Furthermore, ensure that the validation sets contain an equal number of machine-text and paraphrases of machine-text, augmenting them with an equal number of the human-paraphrases we generated.
We follow the same procedure for test set, while additionally mensuring that we have exactly $500$ samples for each category: human-text, paraphrases of human-text, and paraphrases of machine-text.
The validation sets are used to train the edit-based detector discussed in~\autoref{sec:detecting_paraphrases}, while the training sets are used to train both our paraphrase inversion and paraphrase detection models.

\section{Ablations}\label{sec:ablation}

\paragraph{How does varying the sampling procedure impact paraphrase inversion?} 
In ~\autoref{table:ablation_inference_temperature} we show the effect that the decoding temperature has in the quality of the inversions generated by our untargeted inversion model. 
We generate $100$ inversions for every paraphrase in our test dataset, and report metrics using the ``max" scoring strategy dicussed in ~\autoref{sec:methods}.
We observe that temperature plays an important role in the quality of the inversions, with values too low or too high significantly degrading the quality of the inversions.
As the temperature increases, the entropy of the distribution approximates that of a uniform distribution, thereby diffusing the style of the inversions.
Conversely, as the temperature decreases, the inversion model becomes over-confident in its predictive distribution, thereby not exploring neighboring tokens and styles.

\begin{table}[h!]
\centering\small
\begin{tabular}{ccc}
\bf Temperature & \bf Style Sim. & \bf BLEU \\\toprule
0.3 & 0.67 & 0.23 \\
0.5 & 0.69 & 0.24 \\
0.6 & 0.70 & 0.25 \\
0.7 & 0.70 & 0.25 \\
0.8 & 0.71 & 0.24 \\
0.9 & 0.71 & 0.23 \\
1.5 & 0.55 & 0.06 \\
\midrule
\end{tabular}
\caption{Effect of the temperature in the quality of the untargeted inversions.}
\label{table:ablation_inference_temperature}
\end{table}

\paragraph{Are paraphrases generated with lower temperature values easier to invert?}

\begin{table}[h!]
\centering\small
\begin{tabular}{ccc}
\bf Training Temperature & \bf Style Sim. & \bf BLEU \\\toprule
0.3 & 0.71 & 0.26 \\
0.5 & 0.70 & 0.25 \\
0.7 & 0.70 & 0.25 \\
\midrule
\end{tabular}
\caption{Effect of training on a paraphrase dataset generated with different temperature values.}
\label{table:ablation_training_temperature}
\end{table}

To answer this question, we re-generate our human-text paraphrase data with lower temperature values, training and testing the untargeted inversion model in matched temperature conditions. 
We report the results in~\autoref{table:ablation_training_temperature}.
We observe that, as the temperature decreases, the similarity metrics improve. 
We attribute this to the LLMs becoming over-confident in their predictive-distribution, thereby generating less diverse data which in turn is easier to invert.

\section{Prompts}\label{sec:prompts}

\subsection{Paraphrasing}\label{sec:prompts_paraphrasing}

When paraphrasing with an instruction-tuned LLM, we use the following prompt:

\begin{tcolorbox}[colback=gray!5!white,colframe=gray!75!black,title=\textbf{Prompt:}]
\tt
Rephrase the following 
passage: <PASSAGE>

Only output the 
rephrased-passage, do not 
include any other details.

Rephrased passage:
\end{tcolorbox}

We also clean out all obvious generation artifacts, keeping only the paraphrased text.

\subsection{Inversion}\label{sec:prompts_inversion}

\subsubsection{Inversion}

\begin{tcolorbox}[colback=gray!5!white,colframe=gray!75!black,title=\textbf{Prompt:}]
\tt

[INST] The following passage is 
a mix of human and machine text, 
recover the original human text: 
\{generation\}
[/INST]\textbackslash n\#\#\#\#\#\textbackslash nOutput: \{original\}
\end{tcolorbox}

\subsection{Prompting Inversion}

\begin{tcolorbox}[colback=gray!5!white,colframe=gray!75!black,title=\textbf{Prompt:}]
\tt

The following passage is a mix 
of human and machine text,
recover the original human text:
\end{tcolorbox}

\subsection{Generating Reddit Responses}\label{sec:prompt_reddit_response}

\begin{tcolorbox}[colback=gray!5!white,colframe=gray!75!black,title=\textbf{Prompt:}]
\tt

Write a response to the 
following Reddit 
comment: {comment}
\end{tcolorbox}

\section{Dataset Statistics}\label{sec:dataset_statistics}

We show the statistics of the \texttt{Reddit}, \texttt{RAID-ArXiv}, and \texttt{RAID-MovieReviews} in~\autoref{table:statistics_human_paraphrase}.

\begin{table}[h!]
\centering\small
\begin{tabular}{cc}
\bf Split & \bf Number of Examples \\\toprule
\rowcolor[gray]{0.85} \multicolumn{2}{c}{Reddit Human-Paraphrase} \\
Train & 204260 \\
Valid & 24549 \\
Test & 2449  \\
\rowcolor[gray]{0.85} \multicolumn{2}{c}{Reddit Machine-Paraphrase} \\
Train & 239710  \\
Valid & 28883 \\
Test & 2854 \\
\rowcolor[gray]{0.85} \multicolumn{2}{c}{Reddit Machine-Text Detection} \\
Test & 1500 \\
\rowcolor[gray]{0.85} \multicolumn{2}{c}{RAID-ArXiv} \\
Train & 48035 \\
Valid & 3798 \\
Test & 1500  \\
\rowcolor[gray]{0.85} \multicolumn{2}{c}{RAID-MovieReviews} \\
Train & 25649 \\
Valid & 1329 \\
Test & 1500  \\
\midrule
\end{tabular}

\caption{Statistics of the \texttt{Reddit}, \texttt{RAID-ArXiv}, and \texttt{RAID-MovieReviews} datasets.}
\label{table:statistics_human_paraphrase}
\end{table}

\section{Training Hyper-Parameters}\label{sec:training_hyperparameters}

We train all our inversion models with the hyper-parameters shown in~\autoref{table:training_hyperparameters}. 
We train all our models on 4 NVIDIA-A100 GPUs.
Each model took at most 10 hours to train.

\begin{table}[h!]
\centering\small
\begin{tabular}{cc}
\bf Hyper-Parameter & \bf Value. \\\toprule
Learning Rate & $2e^{-5}$ \\
Number of Epochs & 4 \\
LoRA-R & 32 \\
LoRA-$\alpha$ & 64 \\
LoRA-Dropout & 0.1 \\
\midrule
\end{tabular}
\caption{Training Hyper-parameters.}
\label{table:training_hyperparameters}
\end{table}

Most of the compute was spent generating the inversions necessary to run all the experiments, which are in the ballpark of 1M total generations.
We used VLLM~\citep{kwon2023efficient} to speed up the inference time.
We estimate an upper bound of around 150 GPU hours to run all experiments.

\end{document}